\documentclass[11pt,a4paper]{article}
\usepackage[hyperref]{acl2021}
\usepackage{times}
\usepackage{latexsym}

\usepackage{graphicx}
\usepackage{xurl}
\usepackage[]{footmisc}

\usepackage{csvsimple,longtable,booktabs,hhline}
\usepackage{microtype}

\aclfinalcopy 

\title{A diachronic evaluation of gender asymmetry in euphemism}

\author{Anna Kapron-King \\
  Department of Computer Science \\
  University of Toronto \\
  \texttt{annakk@cs.toronto.edu} \\\And
  Yang Xu \\
  Department of Computer Science \\
  Cognitive Science Program \\
  University of Toronto \\
  \texttt{yangxu@cs.toronto.edu} \\}

\date{}

\begin{document}
\maketitle
\begin{abstract}
The use of euphemisms is a known driver of language change. It has been proposed that women use euphemisms more than men. Although there have been several studies investigating gender differences in language, the claim about euphemism usage has not been tested comprehensively through time. If women do use euphemisms more, this could mean that women also lead the formation of new euphemisms and language change over time. Using four large diachronic text corpora of English, we evaluate the claim that women use euphemisms more than men through a quantitative analysis. We assembled a list of 106 euphemism-taboo pairs to analyze their relative use through time by each gender in the  corpora. Contrary to the existing belief, our results show that women do not use euphemisms with a higher proportion than men. We repeated the analysis using different subsets of the euphemism-taboo pairs list and found that our result was robust. Our study indicates that in a broad range of settings involving both speech and writing, and with varying degrees of formality, women do not use or form euphemisms more than men.
\end{abstract}

\section{Introduction}
What role does gender play in language change and use? This question has long been a matter of discussion among linguists. In Robin Lakoff's influential work on this topic, she proposes many ways in which language spoken by women and about women differs from language used by and about men \citep{Lakoff_1973,Lakoff_1977}. Lakoff discusses the causes of these differences and what they tell us about women's role in society. The difference we focus on in this study is one she mentions
only briefly, namely that women use euphemisms more than men do \cite[p.~78]{Lakoff_1977}. Lakoff is not the first to propose this difference. \citet[originally published 1922]{Jespersen_2013} also claims that women use euphemisms more often, and discusses this supposed characteristic of women's language at length. Both Lakoff and Jespersen believe that women use euphemisms more out of a desire to speak more tactfully and to avoid directly mentioning ``unladylike'' topics. For example, Jespersen states that women have often invented euphemisms to avoid mentioning ``certain parts of the human body and certain natural functions'' (\citeyear{Jespersen_2013}, p. 245).

Euphemisms have been considered as an important driver of language change \citep{Burridge_2012}. As a euphemism becomes conventionalized, it may become taboo by its association with a taboo topic, and thus ends up being replaced by a new euphemism. This process has been dubbed the \textit{euphemism treadmill} by \citet{Pinker_1994}. While \citet{Lakoff_1977} considers euphemism use to be a sign of linguistic conservatism, given what we know about the euphemism treadmill it may also be reasonable to associate euphemisms with linguistic innovation. 
Since women are thought to use euphemisms and invent new euphemisms in order to avoid taboo, a finding that women do in fact use euphemisms more could be an indication that women are leading the euphemism treadmill process. 

One may take for granted that women use euphemisms more than men, because this idea has been proposed by two renowned linguistics, and with a few decades between them.
However, both Jespersen and Lakoff base this claim primarily on anecdotal evidence. 
To our knowledge, no one has attempted to quantitatively evaluate whether greater use of euphemisms is characteristic of women's speech. Here we analyze euphemism usage through time by men and women in four large text corpora of English to test this claim. Specifically, we examine whether at a given time point in history, women use euphemisms with a greater proportion in usage frequency than men do. 

\section{Related work}
\subsection{Defining euphemism}

Definitions of euphemism can vary \citep{CasasGomez_2009}. The Oxford English Dictionary (OED) defines a euphemism as ``a less distasteful word or phrase used as a substitute for something harsher or more offensive" \citep{oed_euphemism}. Euphemisms are a common part of everyday polite speech. For example, we might describe certain bodily functions as ``going to the bathroom" or when referring to someone who has just died we might say they have ``recently passed". Euphemisms can be used to discuss any taboo topic, without directly naming the taboo. Taboo is culturally and contextually dependent, and as such so is euphemism \citep{Allan_Burridge_2006, Burridge_2012}.

While there is variation in what different people consider to be a euphemism in different contexts, people can judge words to be euphemistic without needing context, such as when they compile euphemism dictionaries \citep{Burridge_2012}. Allan and Burridge consider these judgements to be made following the ``middle class politeness criterion" (MCPC) \cite[p. 33]{Allan_Burridge_2006}, which vaguely describes a polite, ``middle-class environment" where a euphemism might be preferred over a more offensive expression. This is roughly the context we assume in this study.

Following the OED definition of euphemism, we consider only words and phrases as euphemisms, though a euphemism could conceivably be any length of utterance. We also assume here that a euphemism is an expression that substitutes for a taboo expression, though some scholars argue that there is not always a direct correspondence between a euphemism expression and a taboo \citep{CasasGomez_2009}.

\subsection{Gender and language}

Throughout many different cultures and languages, it has been observed that men and women use language differently, and some of these differences have been remarked upon for centuries \citep{Lakoff_1973,Jespersen_2013, Eckert_McConnell-Ginet_2013, Holmes_1997, Coates_2015, Labov_1994}. This conversation surrounding gender and language has historically often been framed as a characterization of ``women's language" \citep{Jespersen_2013, Lakoff_1973}. Gender differences have been proposed in a wide range of speech characteristics, including word choice, sentence structure, topic choice, and utterance length \citep{Newman_Groom_Handelman_Pennebaker_2008}. In particular, \citet{Lakoff_1973} also proposed that women's speech is more ``polite" than men's, and this has been discussed and studied extensively \citep{Holmes_1997, Brown_1980, Newman_Groom_Handelman_Pennebaker_2008}.

There have been many empirical studies of gender differences in language. Newman and colleagues (\citeyear{Newman_Groom_Handelman_Pennebaker_2008}) used text samples to comprehensively investigate a large number of proposed gender differences in language use. Their results did show evidence for some of these differences, but with small effect sizes. Among some of their findings which supported existing claims were that women use pronouns more than men, that men swear more than women, and that women use polite forms (e.g., ``Would you mind...'') more than men. \citet{Newman_Groom_Handelman_Pennebaker_2008} did not investigate gender differences in euphemism use, but since euphemisms are often considered a form of polite speech \citep{Allan_Burridge_2006, Burridge_2012}, their positive finding that women use polite forms more than men may lend credence to the idea that women use euphemisms more.

More recently, Park and colleagues (\citeyear{Park_et_al_2016}) studied the differences in topics discussed by men and women on Facebook, and how these topics aligned with the interpersonal dimensions of affiliation and assertiveness. They found that women did not use more indirect language than men, contrary to the stereotype that women are less assertive than men and contrary to some of Lakoff's (\citeyear{Lakoff_1973})  claims about women's language. Since euphemisms are a form of indirect speech \citep{Allan_Burridge_2006, Burridge_2012}, this result could be seen to provide evidence against the claim that women use euphemisms more than men.

\subsection{Quantitative approaches to lexical semantic change and euphemism}

There has been much interest recently in the field of computational linguistics and natural language processing in applying quantitative methods to historical language change,
particularly semantic change~\cite{tahmasebi2018survey}. Existing work has explored aspects including but not restricted to the automatic detection~\cite{sagi2011tracing,cook2010automatically,kulkarni2015statistically,schlechtweg2020semeval}, laws~\cite{xu2015computational,hamilton-etal-2016-diachronic,dubossarsky2017outta}, and modeling~\cite{frermann2016bayesian,bamler2017dynamic,giulianelli-etal-2020-analysing} of semantic change. Differing from this line of work, our focus here is to understand the formation and use of euphemism as a driver of language change. To our knowledge, the closest quantitative approaches to euphemism sought to automatically detect euphemism for content moderation~\cite{zhu2021self} and to classify phrases as euphemistic or dysphemistic using sentiment analysis~\cite{felt-riloff-2020-recognizing}, but there exists no quantitative work on characterizing the role of gender in euphemism in a diachronic setting. 

We utilize a set of 106 euphemism-taboo pairs and four large diachronic corpora to test whether women use euphemisms with a higher proportion than men. To verify the robustness of our results, we run the analysis on different subsets of euphemism-taboo pairs to mitigate potential issues with our selection of pairs. Throughout these analyses, we find no evidence that women use euphemisms more than men over time.

In the following, we first describe the quantitative methodology we use to investigate the claim that women use euphemisms more than men, and we then discuss the results.

\section{Methodology}

We quantify euphemism usage by a proportion measure specifying how frequently a given euphemism is used in natural language out of the sum of usage counts of that euphemism and its corresponding taboo expression. This is how we interpret Lakoff and Jespersen's claim that women use euphemisms more. If they only meant that women use euphemisms more without a higher euphemism proportion, then their claim would be simply that women discuss taboo topics more frequently than men, euphemistically or not, which we do not believe is their intention. Hence, we evaluate whether women use euphemisms more by testing whether they tend to have a higher euphemism proportion.

\subsection{Diachronic text corpora}
We analyzed four large diachronic text corpora covering different time periods. We required corpora for which the author or speaker's gender could be determined for each data point. We chose to use longitudinal corpora because euphemisms are known to change over time, and can often be short-lived \cite{Burridge_2012}, and as such we might expect the usage of a given euphemism to change over time. 

The corpora are: Reddit\footnote{\url{https://github.com/ellarabi/gender-idiomatic-language}}, New York Times Annotated Corpus (NYT)\footnote{Only available with license.}, Canadian Parliamentary dataset (Canadian Parl.)\footnote{\url{http://lipad.ca}}, and United States Congressional dataset (US Congr.)\footnote{\url{https://data.stanford.edu/congress_text}} \cite{rabinovich2020pick, nyt_2008,lipad_2017,heinbound_2018}. A summary of statistics for these corpora is shown in Table~\ref{tab:corpora-stats}. These corpora represent a variety of registers; two of the corpora are spoken, three are formal, and one is informal from social media. The NYT, Canadian Parl. and US Congr. likely embody the MCPC context described by \citet{Allan_Burridge_2006}, which makes them good candidates for analyzing euphemism usage. Reddit is a less controlled context, so it may not qualify for the MCPC, which makes it a good point of comparison for the other three corpora.

\begin{table*}[!h]
\centering
\begin{tabular}{ccccc}
\hline
& \textbf{Reddit} & \textbf{NYT} & \textbf{Canadian Parliament} & \textbf{US Congress} \\ \hline
timespan & 2006--2020 & 1987--2007 & 1951--2018 & 1951--2010 \\ 
mean entries per year & 8,138,844 & 88,365 & 40,756 & 138,105 \\ 
initial \% entries by women & 28\% & 9\% & 0.4\% & 0.6\% \\ 
final \% entries by women & 39\% & 21\% & 24\% & 41\% \\ \hline
\end{tabular}
\caption{\label{tab:corpora-stats} A summary of basic statistics for each corpus used in this study. The Canadian Parliament and US Congress datasets are available for earlier years, but the year 1951 was chosen as their starting point because prior to that year the data for women is too sparse.}
\end{table*}

We did not analyze more dated historical corpora for a few reasons. It would be very difficult for us to judge what should be considered a euphemism 100 or more years ago. We would also need data with a high enough proportion of women authors such that we would not have data sparsity issues, and we expect more recent datasets to have larger proportions of women. The article in which Lakoff says women use euphemisms more was published in 1977 \cite{Lakoff_1977}, which falls within the time span of our analysis.

The US Congr., Canadian Parl., and NYT corpora have a large gender imbalance, with only a small (but increasing) percentage of data each year having been produced by women. We perform the analysis of the US Congr. and Canadian Parl. data beginning in 1951, because from this year on the number of speeches by women per year exceeds our chosen sample size of 100, as we describe later.

\subsection{Euphemism-taboo pairs}
In order to analyze the usage of euphemisms compared to taboo expressions on a large scale, we need a data source which pairs euphemism expressions with their equivalent taboos. For example:
\begin{itemize}
    \item passed away (euphemism) $\rightarrow$ died (taboo)
    \item bust (euphemism) $\rightarrow$ breast (taboo)
\end{itemize}
While there are many euphemism dictionaries \cite{Neaman_Silver_1995, Rawson_1981}, and the online OED has a ``euphemism" category by which to browse dictionary entries, the entries in these references do not provide direct correspondences between euphemisms and taboo expressions. They also tend to include antiquated, overly specialized, and highly polysemous euphemisms. For example, Neaman and Silver include \textit{a green hornet} as a euphemism for a motorcycle traffic policeman in Toronto, Canada \cite[p.195]{Neaman_Silver_1995}. To our knowledge, there is no existing list of euphemisms paired with taboo expressions. Our contribution includes a list of 106 pairs of euphemism expressions and taboo expressions, and we have tried by our best judgment to choose expressions found in North American English that are not overly ambiguous or esoteric. 
Table \ref{tab:pairs-tab} shows a subset of the pairs that we analyzed. The complete list is available here: 

\url{https://github.com/annakin6/euphemism-gender}

Some of the pairs come from articles which discuss a perceived societal preference for one phrase over another \cite{Collier_2010, Hayes-Bautista_Chapa_1987, Martin_1991, Nowrasteh_2017, O'Conner_Kellerman_2012, Sagi_Gann_Matlock_2015, Silver_2015, Woelfel_2019, Yandell_2015}, while others we found in euphemism dictionaries and the OED \cite{Neaman_Silver_1995, OED, Rawson_1981}. A remaining minority of pairs were determined from our own knowledge of euphemisms. The euphemisms were chosen to represent a variety of topics, such as illness, body parts, and war. Previous work has shown that men and women tend to discuss different topics \cite{Newman_Groom_Handelman_Pennebaker_2008, Park_et_al_2016}, so it was important to choose topics that would not favour only men or only women. The same expression sometimes appears in this list as both a taboo and a euphemism (in different pairs). This is because we have included pairs that represent different stages of the euphemism treadmill.

\begin{table*}[ht!]
\centering
\begin{tabular}{lll}
    \hline\bfseries euphemism &\bfseries taboo &\bfseries source \\ \hline
\begin{filecontents*}{short_pairs_for_si.csv}
euphemism,taboo,source,topic
african american,black person,\cite{Martin_1991},race
climate change,global warming,\cite{Sagi_Gann_Matlock_2015},politics
custodian,janitor,\cite{Rawson_1981},misc taboo
developing country,third world country,\cite{Silver_2015},politics
handicapped,crippled,\cite{O'Conner_Kellerman_2012},health
homemaker,housewife,\cite{Rawson_1981},gender
illegal immigrant,illegal alien,\cite{Nowrasteh_2017},politics
income inequality,poverty,--,poverty
indigenous,native,\cite{Yandell_2015},race
latino,hispanic,\cite{Hayes-Bautista_Chapa_1987},race
overweight,obese,\cite{Collier_2010},health
same-sex,gay,--,sexuality
underpriviledged,poor,\cite{Rawson_1981},poverty
undocumented immigrant,illegal immigrant,\cite{Nowrasteh_2017},politics
\end{filecontents*}
\csvreader[late after line=\\, late after last line=\\\hline
]{short_pairs_for_si.csv}{1=\euphemism,2=\taboo,3=\source}{\euphemism & \taboo & \source}
    
\end{tabular}
\caption{A subset of the lemmatized euphemism-taboo pairs used in this study, with their corresponding sources (or -- if no source could be identified). \label{tab:pairs-tab}}
\end{table*}

\subsection{Quantification of euphemism usage proportion}

To determine euphemism use by gender, we first divided the data according to the speaker's gender. The Reddit data was already separated into self-reported binary gender categories, but the other three corpora did not explicitly contain this information (except for some rare entries in the US Congr. dataset). To classify gender, we first used the speaker's title (e.g., Mr.) if it was clearly masculine or feminine. If the title could not be used, we relied on the R \texttt{gender} package to determine gender from the speaker's first name. This package allows for gender retrieval given a first name and a birth year range. Using this package, we created 40-year bins for every decade from 1930 to 1990, and we considered the birth year of a given author/speaker to between 20 to 40 years before the decade that their article/speech was produced. For example, the gender of a speaker from 1951 would be determined from classification data for the birth years 1890--1930. Any texts for which a binary gender classification could not be determined were discarded. For this reason, 17\% of the Canadian Parl. data was discarded, 0.04\% of the US Congr. data, and 46\% of the NYT data.

After dividing by gender, we selected a random sub-sample of fixed size from each gender (100 speeches for Canadian Parl. and US Congr., 1000 articles/posts for NYT and Reddit), to make up for the gender imbalance. For each euphemism-taboo phrase pair we counted the number of times the euphemism and the taboo expression occur in the sample. For each corpus and each pair, we computed a euphemism usage proportion \textit{p} for each gender, as shown in Equation~\ref{eq:ratio}, where $f_{y}^{g}$ is the frequency of the expression in the sample for gender \textit{g} and year \textit{y}. 
\begin{eqnarray}
\label{eq:ratio}
\resizebox{.9\hsize}{!}{$
    p(\mathrm{euphemism}, \mathrm{taboo}, g, y) =  \frac{f_{y}^{g}(\mathrm{euphemism})}{f_{y}^{g}(\mathrm{euphemism}) + f_{y}^{g}(\mathrm{taboo})}
    $}
\end{eqnarray}

To ensure plural, past tense, and other inflected forms of the euphemisms are not overlooked, both the corpora and the euphemism-taboo pairs were lemmatized using the \texttt{nltk} \texttt{WordNetLemmatizer} and part of speech tagger. This means that each word is reduced to its base lemma, as informed by its part of speech (POS) tag. For example, the word \textit{women}, when its POS is noun, becomes \textit{woman}, and the word \textit{deprived}, when its POS is adjective, remains \textit{deprived}. Lemmatization was used as it provided a more comprehensive and accurate collection of euphemisms in various forms than a stemmer, which just removes affixes, would. For example, the \texttt{PorterStemmer} would return \textit{women} and \textit{depriv}, irrespective of POS.

\section{\label{section:results}Results}
We first analyze for all euphemism-taboo pairs, the number of times they are used by each gender, and for how many of these pairs women have a higher euphemism proportion than men do. We then perform focused analyses on selected subsets of the list of euphemism-taboo pairs. These analyses respectively examine: only pairs where the euphemism or taboo is a multi-word phrase, and only pairs which meet the cut-off threshold for all four corpora (i.e., the conjunction set of euphemism-taboo pairs). We find in all the analyses that women do not use euphemisms more than men.

\begin{figure*}[ht!]
\centering
\includegraphics[scale=0.385]{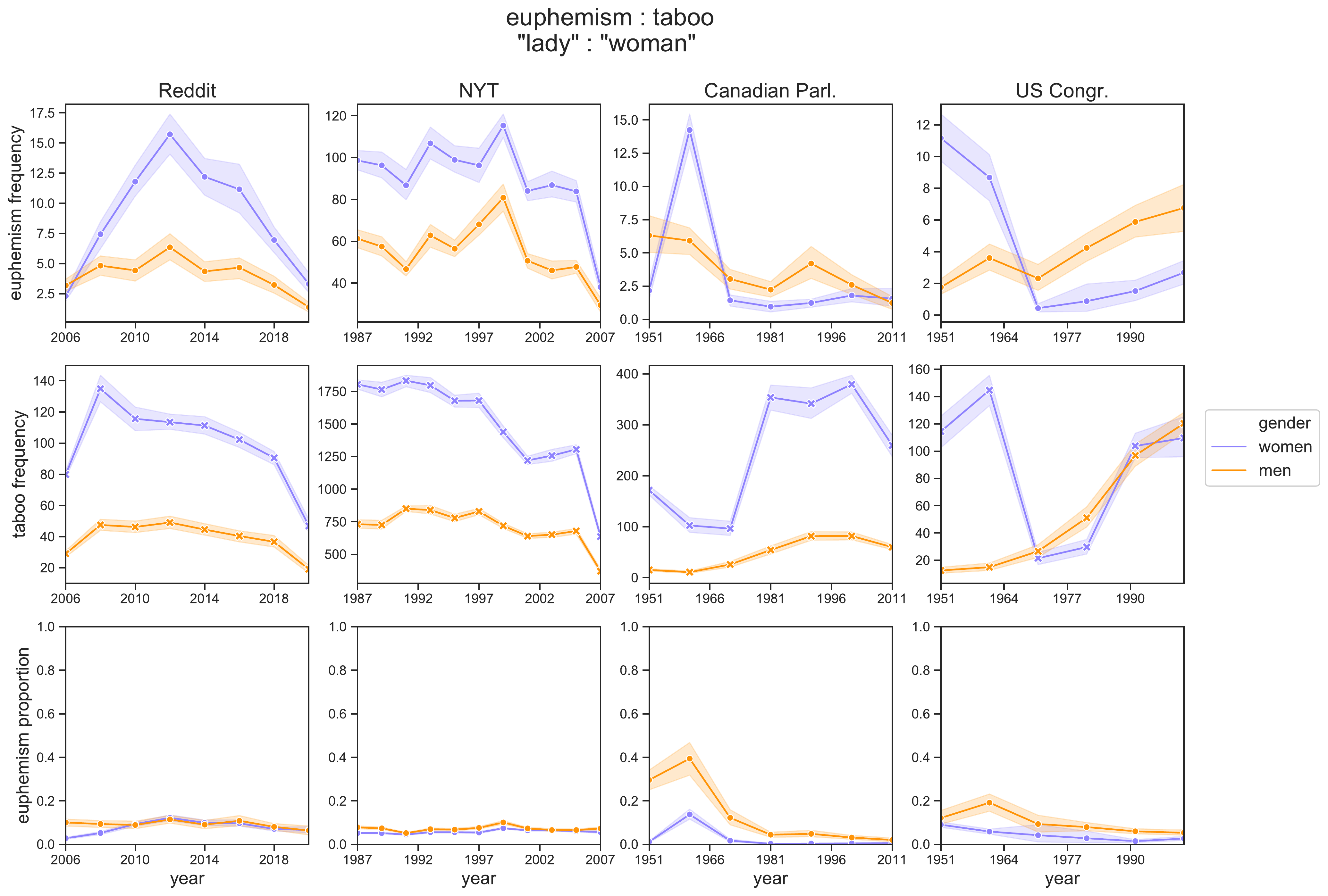}
\caption{Euphemism and taboo frequencies over time. The top row shows raw euphemism frequency over time, averaged over samples and time spans (span = 2 years for Reddit and NYT, 10 years for Canadian Parl. and US Congr.). The shaded area indicates a 95\% confidence interval. The second row shows raw taboo frequency over time, averaged over samples and time spans. The y-axes for the first and second row are not fixed to the same scale, since they are only meant to illustrate the relative difference in frequency for men and women. The third row shows euphemism proportion over time [0,1], as defined in Equation~\ref{eq:ratio}. The axis indicating year differs depending on the timescale of the dataset.}
\label{fig:ladyVwoman}
\end{figure*}

To consider statistical variation in the analyses, we repeatedly sampled 25 times for each gender at each year for each corpus. To alleviate data sparsity, we placed the counts for the US Congr. and Canadian Parl. in 10-year bins, and placed the data for Reddit and NYT in 2-year bins. We also excluded all pairs which did not meet a certain frequency threshold, to eliminate very sparse, unreliable results. For each euphemism-taboo pair and each corpus, we check that both the euphemism and the taboo appear at least once in 10\% or more of the 25 samples. If not, we omit that pair from the results for that corpus. The number of pairs out of the 106 that meet the frequency threshold for Reddit is 32, NYT is 80, Canadian Parl. is 35, and US Congr. is 34.

Using a one-tailed Welch's unequal variances t-test, for each pair we compared the euphemism proportions in all 25 samples between genders. We recorded the fraction of pairs for which women had a significantly higher euphemism proportion than men at $p<0.05$, and vice versa for men. If women use euphemisms more than men, we would expect this to return a large percent of pairs where woman have a higher euphemism proportion than men, and a smaller percent of pairs for men.

Figure~\ref{fig:ladyVwoman} shows the euphemism and taboo expression frequencies by gender over time for the pair \textit{lady} (euphemism) vs. \textit{woman}. The top two rows show binned frequencies, and the third row shows the euphemism proportion from Equation~\ref{eq:ratio}. This pair is one that specifically relates to women, and that \citet{Lakoff_1973} proposed and explained in detail. We see as expected that women say both \textit{lady} and \textit{woman} more than men do. However, women do not say the euphemism \textit{lady} with a higher proportion than men do. In fact, we find that men use this particular euphemism with a substantially higher proportion than women in three of the four corpora for almost their entire time spans. 

\begin{figure*}[ht]
\centering
\includegraphics[scale=.53]{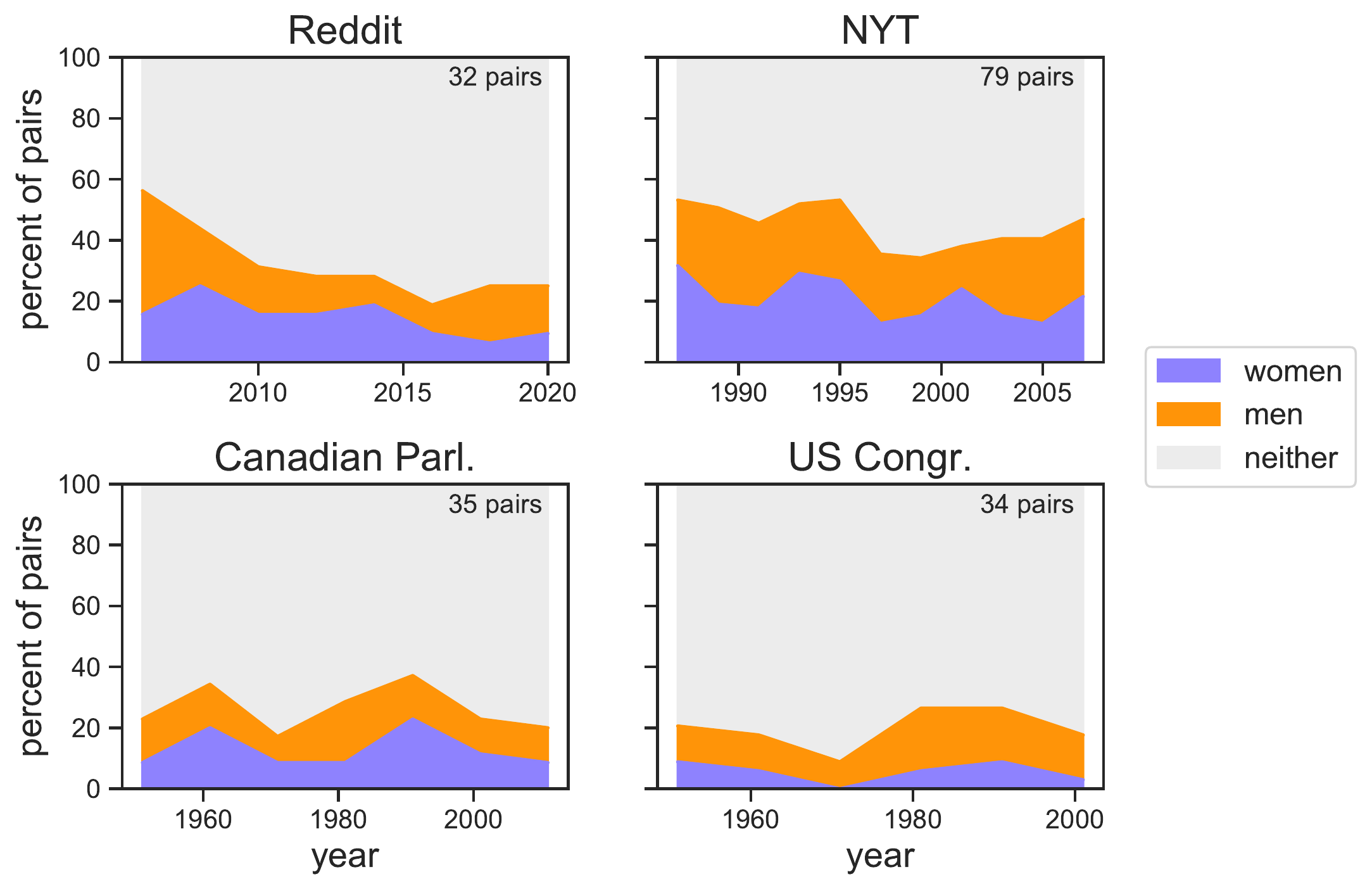}
\caption{Percentage share of pairs (out of all pairs that met the frequency threshold for a given corpus) for which women have a significantly higher euphemism proportion of usage frequency than men, men have a significantly higher euphemism proportion than women, or neither gender has a significantly higher euphemism proportion.}
\label{fig:sig_percents}
\end{figure*}

To summarize the euphemism proportion results, Figure~\ref{fig:sig_percents} shows the percent of pairs over time where either women have a significantly higher euphemism proportion than men, men have a significantly higher euphemism proportion than women, or neither gender has a significantly higher euphemism proportion. 

The individual euphemism-taboo pair plots and the euphemism proportion summary plots show that in all corpora, across the entire time span of 1951--2018, women do not lead in their euphemism-taboo usage proportion. The majority of pairs show no clear leader between men and women. There are some euphemism-taboo pairs where both expressions are said more frequently by women than men, for example women say \textit{lady} and \textit{woman} far more than men do in all four corpora, which is to be expected.

We also repeated this analysis with bins three times larger for each corpus to see if data sparsity affected our results (6 years for Reddit and NYT, 30 years for Canadian Parl. and US Congr.). This increases the number of euphemism-taboo pairs which surpass the frequency threshold for all four corpora, but we found the results still hold that women do not have a consistently higher euphemism proportion than men, and that most pairs show no significant difference in euphemism proportion. 

\begin{figure*}[ht]
\centering
\includegraphics[scale=0.53]{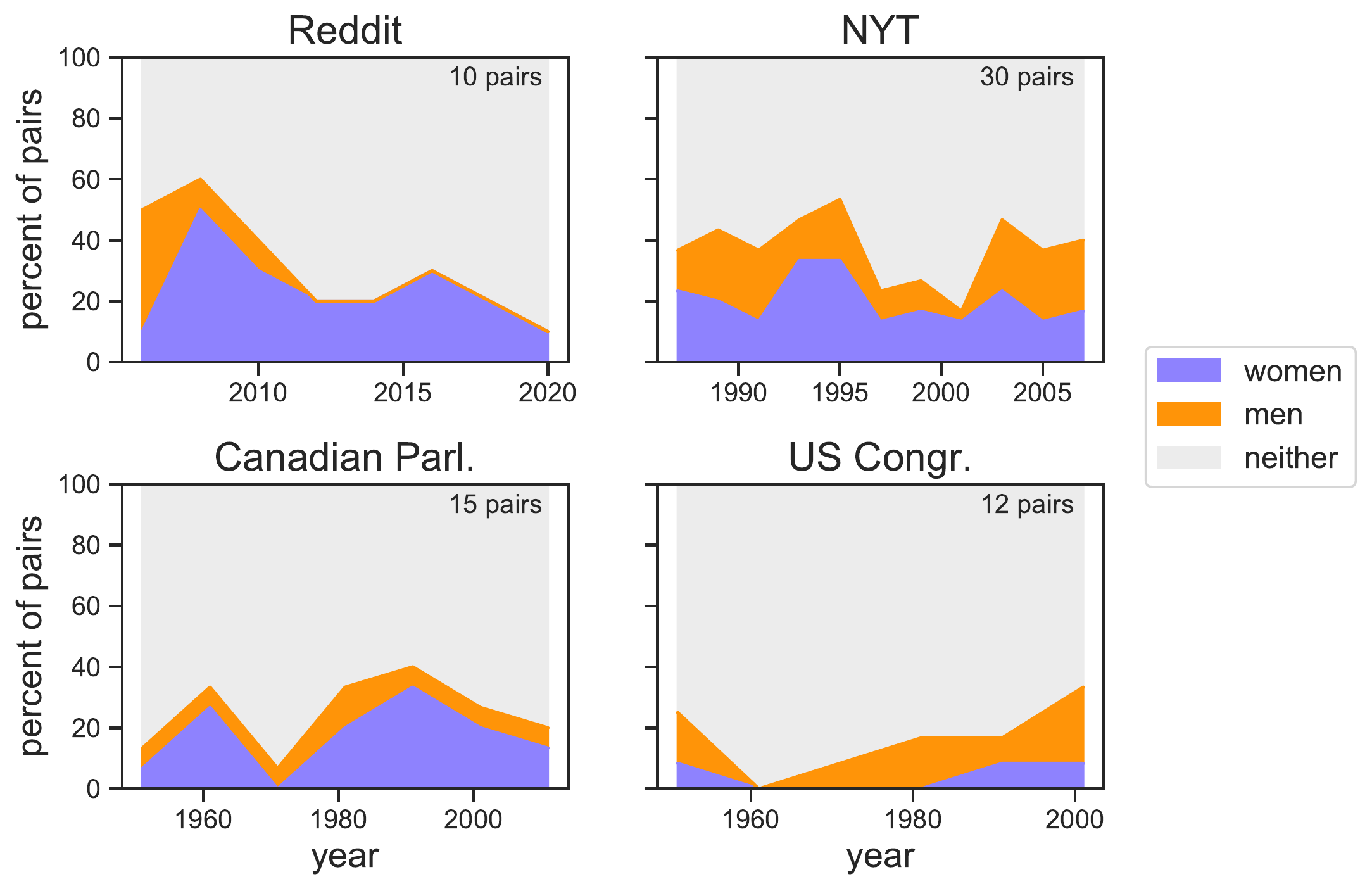}
\caption{Percent of significant euphemism proportion in the phrase-based analysis.
Same as Figure~\ref{fig:sig_percents} but considering only the euphemism-taboo pairs where at least one of the two expressions is a multi-word phrase.}
\label{fig:sig_percent_phrases_only}
\end{figure*}

Although we tried not to include many ambiguous expressions, some of the words in our euphemism-taboo pair list do contain multiple senses. For example, \textit{weed} could be euphemism for \textit{marijuana} or it could refer to an unwanted plant in a garden. Since multi-word phrases are less likely to have multiple senses, we ran the same analysis on only those euphemism-taboo pairs which contain a multi-word phrase. For example, the pair \textit{armed conflict} and \textit{war}. This analysis does not completely handle ambiguity, since multi-word phrases can be ambiguous and we still permit one of the expressions in the pair to be a single word, however it does help mitigate the effect of expressions with alternative non-euphemistic and non-taboo senses on our results. The results, shown in Figure~\ref{fig:sig_percent_phrases_only}, generally support our finding from the complete analysis that there is a minimal difference in how much men and women use euphemisms. The Reddit and Canadian Parliament graphs show women using euphemisms more, but due to the small sample size (10 pairs and 15 pairs respectively) this result is not very reliable.

\begin{figure*}[!h]
\centering
\includegraphics[scale=0.53]{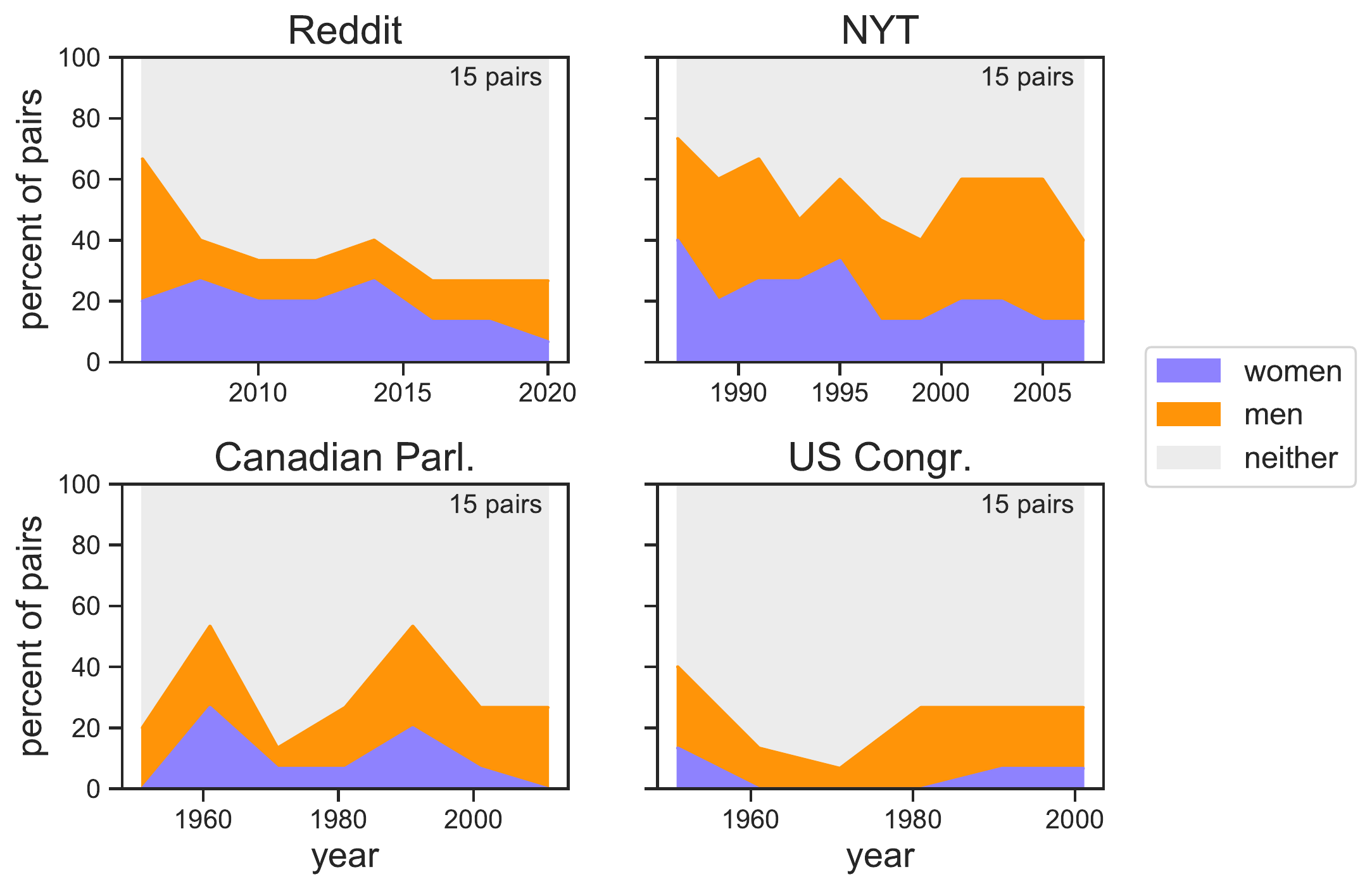}
\caption{Percent of significant euphemism proportion in the conjunctive pairs.
Same as Figure~\ref{fig:sig_percents} but considering only the euphemism-taboo pairs which meet the threshold for all of the four corpora.}
\label{fig:sig_percent_conjunction}
\end{figure*}

\begin{figure*}[h!]
\centering
\includegraphics[scale=0.50]{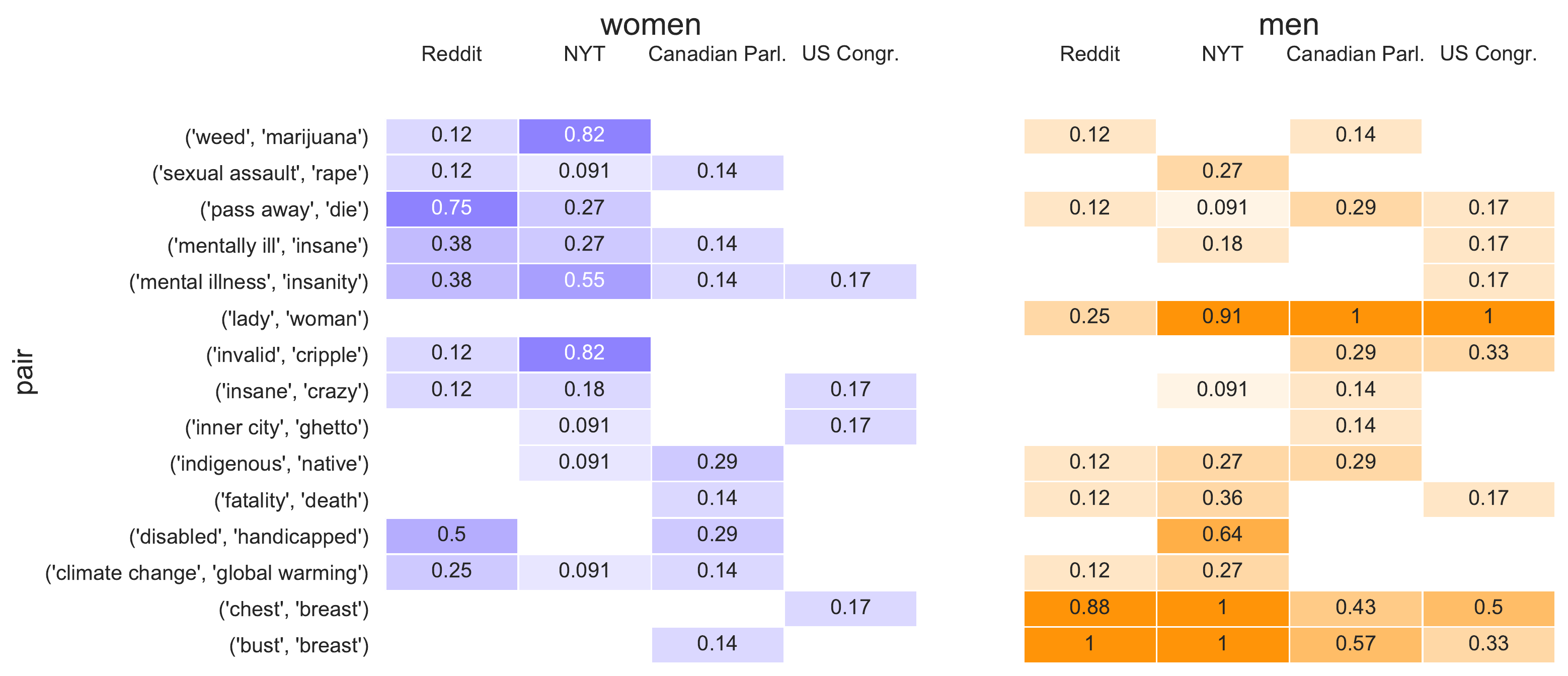}
\caption{Percent of time with significantly higher euphemism proportion (using the same bins we have used throughout, as described in Section~\ref{section:results}).
The left heatmap indicates for a given (euphemism, taboo) pair for what percent of time women had a significantly higher euphemism proportion for that pair than men did. Darker purple blocks were higher for more time, empty blocks never had a significantly higher euphemism proportion for women. The right heatmap shows the same for men, with darker orange indicating a larger percent of time.}
\label{fig:heatmap}
\end{figure*}

For the final analysis, we examine only the pairs which pass the sparsity frequency threshold for every corpus. There are 15 such euphemism-taboo pairs used in this analysis, the results for which are shown in Figure~\ref{fig:sig_percent_conjunction}. 
We can also visualize the amount of time for which each gender exhibits a significantly higher euphemism proportion for 
each pair as a heatmap, shown in Figure~\ref{fig:heatmap}. The heatmap shows that there are no pairs for which women consistently have a higher euphemism proportion than men for a larger period of time across all four corpora. However, there are some pairs in some corpora that stick out. The large percent of time for which women have a higher proportion of saying \textit{weed} compared to \textit{marijuana} than men do in the NYT dataset could likely be explained by the fact that \textit{weed} is one of the more ambiguous words in our set of 106 pairs and may be more commonly associated with gardening than marijuana.

These results again support our finding that women do not use euphemisms more than men. There are not many pairs where women have a higher euphemism proportion for very long, and the pairs for which they do are not consistent across corpora.
There are, however, a few pairs where men consistently have a higher euphemism proportion than women. For the pair \textit{lady}-\textit{woman} and the two pairs containing \textit{breast}, men prefer to use the euphemism more than women do, across all four corpora.

\section{Discussion}
Our analysis of four large, varied datasets spanning 1951--2018 provides no support for Lakoff and Jespersen's claim that women use euphemisms more than men do. This result means that we cannot assume that women use euphemisms more. In general, we should not take for granted the characterization of women's language that has been proposed by linguists such as Jespersen and Lakoff, as our study and others have shown that they are not always supported by empirical evidence \cite{Newman_Groom_Handelman_Pennebaker_2008, Park_et_al_2016}.

The result that women do not use euphemisms more indicates that euphemisms should not be lumped in with other polite forms that women were found to use more in other studies \cite{Newman_Groom_Handelman_Pennebaker_2008}. Our result is consistent with Park et al.'s (\citeyear{Park_et_al_2016}) finding that women do not tend to use more indirect language than men. 

Our finding raises the question, why do Jespersen and Lakoff say that women use euphemisms more if this is not actually the case? One reason women might appear to use euphemisms more is that they may also talk more about certain taboo topics. For example, in our results women say both \textit{chest} and \textit{breast} more than men do, but men say \textit{chest} with a higher proportion than women. 
The topic of the euphemism seems to have some effect on whether men or women use it more, though at a glance these do not seem to correspond to topics that have been found to be discussed more by one gender or the other. There are of course euphemisms that women do prefer to use more than men do, at least within a certain context, such as women on Reddit preferring to say \textit{pass away} over \textit{die} more than men. The claims made by Jespersen and Lakoff could be due to generalizing from specific cases similar to this one. It is also likely that gender differences in language have changed since Jespersen's time -- we were unable to investigate this due to data sparsity and a lack of sources for euphemisms from the 1920s. Even so, our corpora spanned 1951--2020 and our finding was consistent throughout that time period.

The four corpora used in this study were chosen because we needed large diachronic corpora for which the author's gender could be approximately determined. However, there are limitations to using these corpora. The language used in political proceedings and in newspapers may be regulated by political parties or the newspaper editors, which might minimize gender differences in language in these corpora. The Reddit data only included posts where users had self-reported gender, which may limit the topics that are included. We did not include any natural conversation data, although that is likely the setting that Lakoff and Jespersen were most concerned with.

There are also limitations to our selected list of euphemism-taboo pairs. The list is relatively small, and was gathered manually. This list does not represent all taboo topics, nor all types of euphemisms. However, we believe it serves as a good first step for quantitative studies of euphemism. Future work on automatic euphemism detection may allow us to generate a more comprehensive list which should  help with analyzing euphemism use over time and other related phenomena.

Our study did not directly examine whether women lead euphemism innovation and change, although our result does indicate that this is not likely to be generally true. Future work could investigate who leads the formation of new euphemisms and who drives the euphemism treadmill, while considering that the answer to this question is likely context- and topic-dependant.

\section{Conclusion}
The subject of how women's language and men's language differ is one that has been extensively discussed, and one alleged difference is that women use euphemisms more than men do. However, this claim has been based on anecdotal evidence. Our diachronic evaluation using large corpora spanning multiple decades from a variety of contexts shows that women do not use euphemisms more than men do. Our work indicates the importance of using quantitative methods to evaluate long-held beliefs about language use and language change.



\newpage

\bibliographystyle{acl_natbib}

\bibliography{anthology,acl2021}

\end{document}